# TOPOGRAPHIC FEATURE EXTRACTION FOR
# BENGALI AND HINDI CHARACTER IMAGES

Soumen Bag and Gaurav Harit

Department of Computer Science and Engineering, Indian Institute of Technology,
Kharagpur, West Bengal-721302, INDIA
`{soumen, gharit}@cse.iitkgp.ernet.in`

*ABSTRACT*

*Feature selection and extraction plays an important role in different classification based problems such as face recognition, signature verification, optical character recognition (OCR) etc. The performance of OCR highly depends on the proper selection and extraction of feature set. In this paper, we present novel features based on the topography of a character as visible from different viewing directions on a 2D plane. By topography of a character we mean the structural features of the strokes and their spatial relations. In this work we develop topographic features of strokes visible with respect to views from different directions (e.g. North, South, East, and West). We consider three types of topographic features: closed region, convexity of strokes, and straight line strokes. These features are represented as a shapebased graph which acts as an invariant feature set for discriminating very similar type characters efficiently. We have tested the proposed method on printed and handwritten Bengali and Hindi character images. Initial results demonstrate the efficacy of our approach.*

## 1. INTRODUCTION

Feature selection and extraction has wide range of application for pattern recognition. It plays an important role in different classification based problems such as face recognition, signature verification, optical character recognition (OCR) etc. From past several decades, many feature selection and extraction methods are reported [1] for Indian character recognition. The accuracy rate of OCR depends on feature sets and classifiers [2] [3]. Next we begin a brief description about few important feature sets used in optical character recognition for Bengali and Hindi documents.

Chaudhuri and Pal [4] have proposed the first complete printed Bengali OCR in 1998. In this method, nine different strokes are used as primary feature set for recognizing basic characters and template-based features are used for recognizing compound characters. To recognize handwritten basic and compound characters, Das *et al*. [5] have used shadow, longest run, and quad-tree based features. Dutta and Chaudhury [6] have proposed topological features, such as junction points, hole, stroke segments, curvature maxima, curvature minima, and inflexion points of character images for performing printed and handwritten Bengali alpha-numeric character recognition. Later, Bhattacharya *et al*. [7] [8] have also used the same topological feature set for handwritten Bengali numerals. To detect convexity of Bengali numerals, Pal and Chaudhuri [9] have used water-flow model. They also used topological and statistical features for preparing feature set. Bhowmik *et al*. [10] have proposed a stroke-based feature set for recognizing Bengali handwritten character images. In this method, ten stroke based features which indicate the shape, size and position information of a digital curve with respect to the character image, are extracted from character images to form the feature vector. Majumdar [11] has introduced a new feature extraction method based on the curvelet transform of morphologically altered versions of an original character image. Table 1 gives a summary of different feature sets used in Bengali OCR systems.

Table 1. Different feature sets used in Bengali OCR systems

| **Method** | **Feature Set** |
|---|---|
| Chaudhuri and pal [4] | Structural and Template |
| Das *et al*. [5] | Shadow, Longest run, and Quad-tree |
| Dutta and Chaudhury [6] | Structural and Topological |
| Bhattacharya *et al*. [7] [8] | Topological and Structural |
| Pal and Chaudhuri [9] | Watershed, Topological, and Statistical |
| Bhowmik *et al*. [10] | Stroke-based |
| Majumder [11] | Curvlet coefficient |

The first complete OCR on Hindi (Devanagari) is introduced by Pal and Chaudhuri [12]. In this method, they have used structural and template features for recognizing basic, modified, and compound characters. To recognize real-life printed documents of varying size and font, Bansal and Sinha [13] have proposed statistical features. Later Pal *et al*. [14] and Kompalli-Setlur [15] have used the gradient features for recognizing handwritten Hindi characters. Bajaj *et al*. [16] have used density, moment of curve and descriptive component for recognizing Hindi handwritten numerals. Sethi and Chatterjee [17] have proposed a set of primitives, such as, global and local horizontal and vertical line segments, right and left slant, and loop for recognizing handwritten Hindi characters. Sharma *et al*. [18] have used directional chain code information of the contour points of the characters for recognizing handwritten Hindi characters. Table 2 gives a summary of different feature sets used in Devanagari OCR systems.

...Signal & Image Processing : An International Journal (SIPIJ) Vol.2, No.2, June 2011

Table 2. Different feature sets used in Hindi OCR systems.

| Method | Feature Set |
|---|---|
| Pal and Chaudhuri [12] | Structural and Template |
| Bansal and Sinha [13] | Statistical |
| Pal *et al*. [14] | Gradient |
| Kompalli and Setlur [15] | Gradient |
| Bajaj *et al*. [16] | Density, Moment of curve, and Descriptive component |
| Sethi and Chatterjee [17] | Watershed, Topological, and Statistical |
| Sharma [18] | Directional chain code information of contour points |

All of the above mentioned methods do not consider shape variation for extracting features. But in Indian languages, a large number of similar shape type characters (basic and conjunct) are present. From that point of view, we have proposed novel features based on the topography of a character to improve the performance of existing OCR in Indian script, mainly for Bengali and Hindi documents. The major features of the proposed method are listed as follows:

 1. The main challenge to design an OCR for Indian script is to handle large scale shape variation among different scripts. Strokes in characters can be decomposed into segments which are straight lines, convexities or closed boundaries (hole). In our work we consider the topography of character strokes from 4 viewing directions. In addition to the different convex shapes formed by the character strokes, we also note the presence of closed region boundaries.

 2. The extracted features are represented by a shape-based graph where each node contains the topographic feature, and they all are placed with respect to their centroids and relative positions in the original character image.

 3. The proposed methodology is applicable for different languages.

 4. This approach is applicable for printed as well as handwritten text documents.

 5. This topographic feature set helps to differentiate very similar type characters (similarity in shape) in a proper way.

This paper is organized as follows. Section 2 describes the proposed topographic features extraction method. Section 3 contains the experimental results. This paper concludes with some remarks on the proposed method and future work in Section 4.

183

Signal & Image Processing : An International Journal (SIPIJ) Vol.2, No.2, June 2011

## 2. PROPOSED TOPOGRAPHIC FEATURE EXTRACTION METHOD

In this section we describe our proposed feature extraction method based on measurement of convexity of character strokes from different viewing directions. Figure 1 shows the system architecture of the proposed method.

### 2.1. Preprocessing

Given a scanned document page we binarize it using Otsu's algorithm [19]. Currently we are working with printed documents with all text content. For Bengali or Hindi documents the entire word gets identified as a single connected component because of the *matra/shiro-rekha* (head line) which connects the individual characters. For this case we separate out the individual *akshara* within a word by using the character segmentation methods reported in [20] (for Bengali) and [21] (for Hindi).

### 2.2. Thinning for Image Skeletonization

Before extracting the features, character images are converted to thinned (i.e., single pixel thick) curve segments. But to retain proper shape of thinned character images (particularly for India languages) is still a big challenge. Traditional thinning strategies lead to deformation of character shapes, especially where the strokes get branched out; i.e. the junction points in the skeleton. Lot of works have been done on thinning, but most of them are on English, Chinese, and Arabic languages [22] [23] [24] [25] [26] [27]. From this point of view, Bag and Harit [28] have proposed an improved medial-axis based thinning strategy for Indian character images. It uses shape characteristics of text to determine the areas within the image region to stop thinning partially. These regions are marked as junction points. This approach helps to preserve the local features and true shape of the character. These resultant skeletons have stronger ability to cope with all shortcomings of existing mask based thinning methods and would offer better models for feature extraction and classification for OCR. Few results on printed Bengali and Hindi characters are shown in Figure 2.

### 2.3. Extraction of Topographic Features

Topographic features are classified into three categories: closed region, convexity of strokes, and straight line strokes. The convexity of curve is detected from different directions. Here we consider four directions (North, South, East, and West) for convexity measurement.

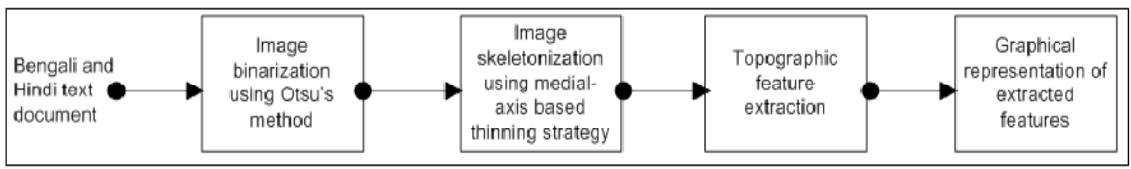

Figure 1. System architecture of the proposed feature extraction method

**Definition of convexity:** The convexity of a shape is detected by checking its convergence towards a single point or a cluster of points, connected by 4-connectivity. For any convexity shape, if we move down word then we shall reach to a single point or a flat region which is a set of 4-connected pixels. For detecting the stroke convexity, we have used this concept. Figure 3 shows the different convexities of character images detected from different view directions. In the figure, different colors are used to mark convexities detected from different view directions (Red, Blue, Green, and Magenta for North, South, East, and West respectively).

The following steps discuss the methodology to detect convexity from North direction.

1. Prepare a database containing different convex shape. Each convex shape is defined according to their structural property. Figure 4 shows different shapes stored in the database.

2. Detect horizontal-start ($H_s$), horizontal-end ($H_e$), vertical-start ($V_s$), and vertical-end ($V_e$) points of the thinned image. Here [$H_s$, $H_e$] and [$V_s$, $V_e$] are the horizontal and vertical limits of the input image respectively (Figure 5).

3. Given a thinned binary image, it is scanned from top to bottom and left to right, and transitions from white (background) to black (foreground) are detected. The whole scanning is done from $H_s$ to $H_e$ and $V_s$ to $V_e$ in horizontal and vertical directions respectively.

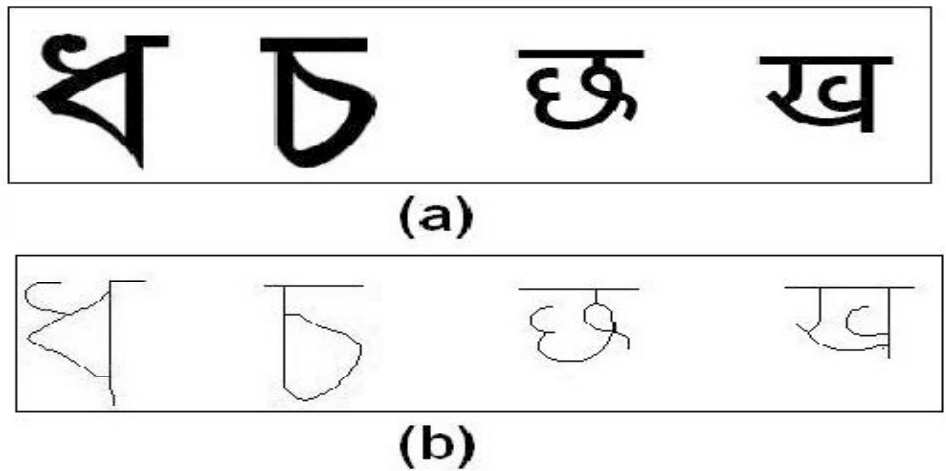

Figure 2. Skeletonization results: (a) Input image; (b) Skeleton image.

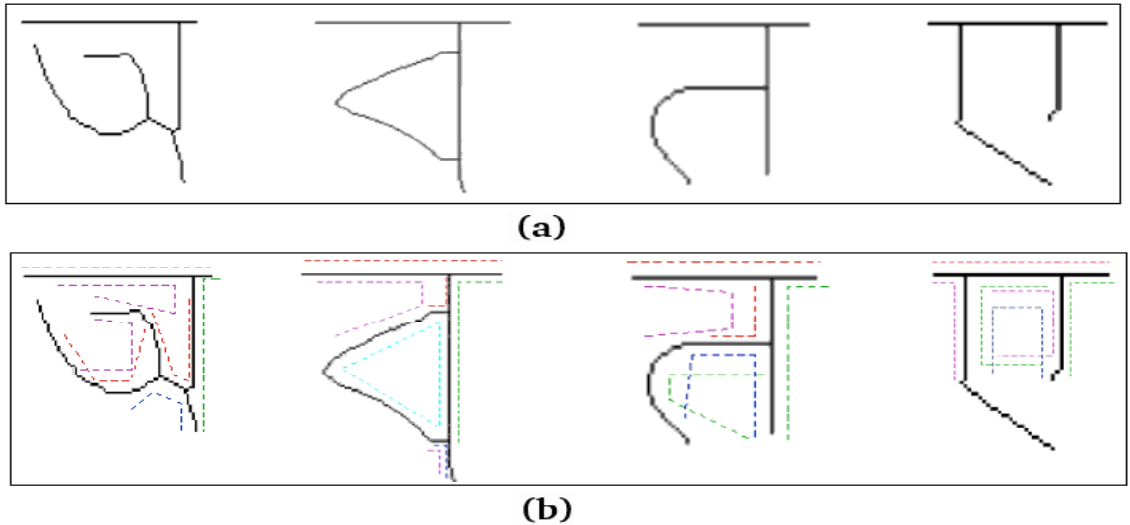

Figure 3. (a) Skeleton image; (b) Different convex shapes marked by different colors (Red, Blue, Green, and Magenta for North, South, East, and West respectively).

    4. Suppose, a single scan from left to right generates a sequence of pixels $\{x_1, x_2, x_3, ..., x_i, x_{i+1}, ..., x_n\}$ where each pixel is a cut point between horizontal scan line and character image. Now, if the two consecutive pixels ($x_i$ & $x_{i+1}$) are not 4-connected neighbors, then put them in an array of points $P[1:N,1:2]$ where each cell $P[i]$ contains one pixel with its $x$, $y$ coordinates in $P[i][1]$ and $P[i][2]$ respectively.

    5. Continue the above steps for all remaining horizontal scan until we get a single point or a set of horizontally 4-connected neighbor pixels with value $\geq \xi$ (set to 5).

    6. The detected set of pixels is matched with the defined convex shapes stored in the database to get a specific convex shape.

    7. The above steps are repeated for detecting the convexity from remaining three directions, i.e. South, East, and West. Only difference is that, for South direction, the scan is done from bottom to top and left to right, and for East/West directions, scan is done from top to bottom and right to left (for East)/left to right (for West) directions. Finally, a set of different convex shapes are generated to prepare topographic feature set.

The above steps are used for detecting convexity of character strokes from different directions. Now for detecting closed region, we use the concept of connected component analysis. If the pixels of the stroke segment are connected (i.e. each pixel has two 8-connected neighbors), then the stroke is identified as **closed region**.

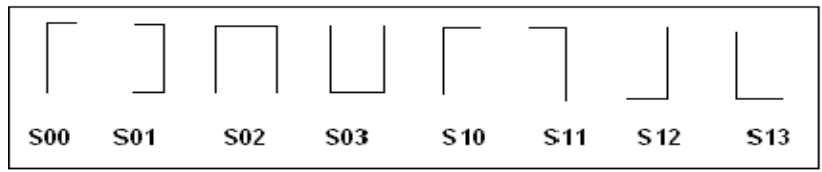

Figure 4. Different convex shapes

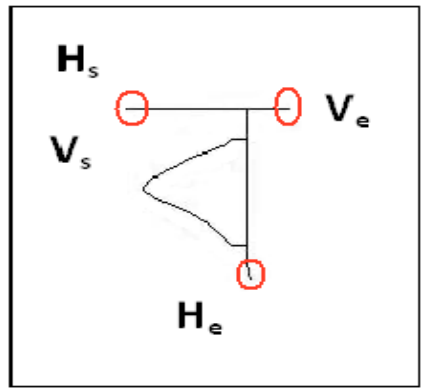

Figure 5. Horizontal and vertical limits of an image

If the number of horizontally 8-connected neighbor pixels is $\geq \epsilon$ (set to 20), then we can say that **straight line** is present. For Bengali and Hindi language, most of the characters have headline. By observing the structural shape of these two types of characters, we can say that the straight line will detect only from North direction. For this reason, we do not use straight detection method for other three directions.

Figure 6 shows the different topographic features extracted from Bengali and Hindi thinned character images.

### 2.4. Graphical Representation of Topographic Features

After detecting the convexity from four directions, we design a shape-based graph to represent the feature set of a particular character. The steps are given below.

1. Suppose, a thinned character image has *k* number of topographic components $\{T_1, T_2, ..., T_k\}$.

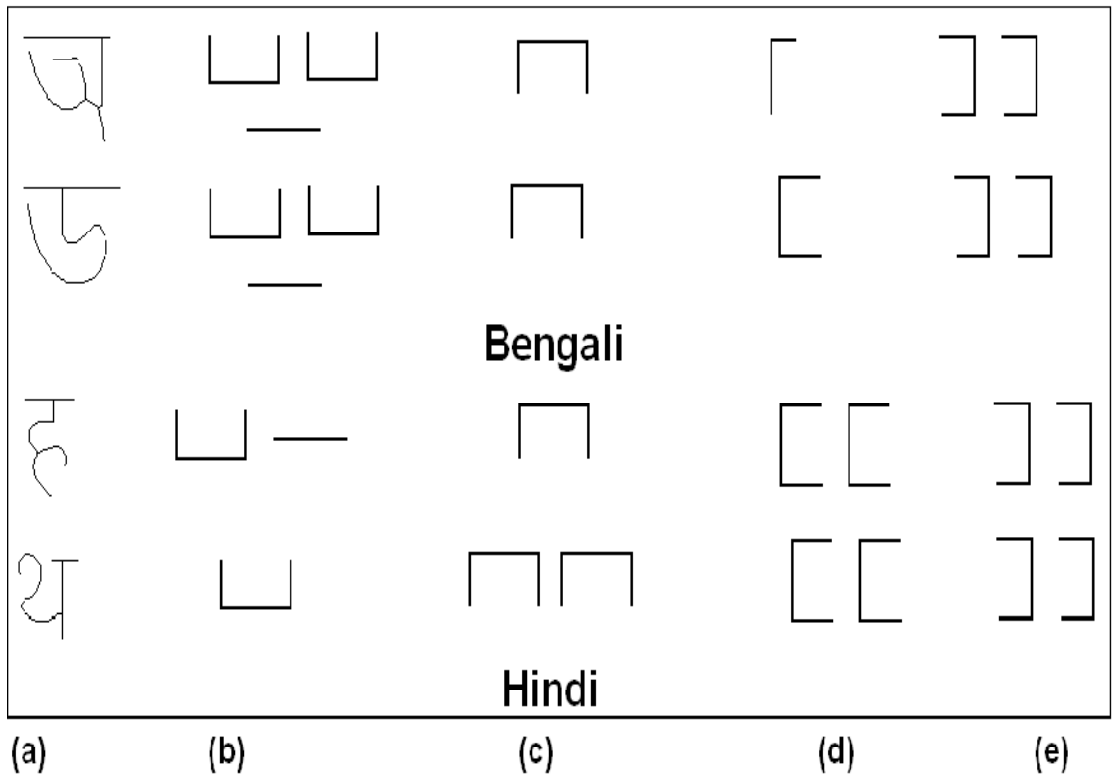

Figure 6. Topographic features of thinned character images: (a) Skeleton image; (b)-(e) Features from North, South, East, and West direction.

2. For each component, calculate the centroid $(X_i, Y_i)$ using the equation given below.

$$X_i = \frac{\sum_{j=1}^{N_i} p_{ij}}{N_i} \; ; \quad Y_i = \frac{\sum_{j=1}^{N_i} p'_{ij}}{N_i}$$

where $N_i$ is the total number of black pixels of the $i^{th}$ topographic component, $p_{ij}$ and $p'_{ij}$ are the x-coordinate and y-coordinate of the $i^{th}$ pixel of the $j^{th}$ topographic component respectively.

3. Design an undirected graph $G = (V, E)$, where $V$ is the set of vertices containing different topographic features and $E$ is the set of edges. All vertices are placed based on the coordinates of there centroid. Now add edges among these vertices with respect to their relative positions in the thinned character image.

Figure 7 shows the shape-based graph of thinned Bengali and Hindi character images. This graphical model gives a clear pictorial difference among different feature set of different character images.

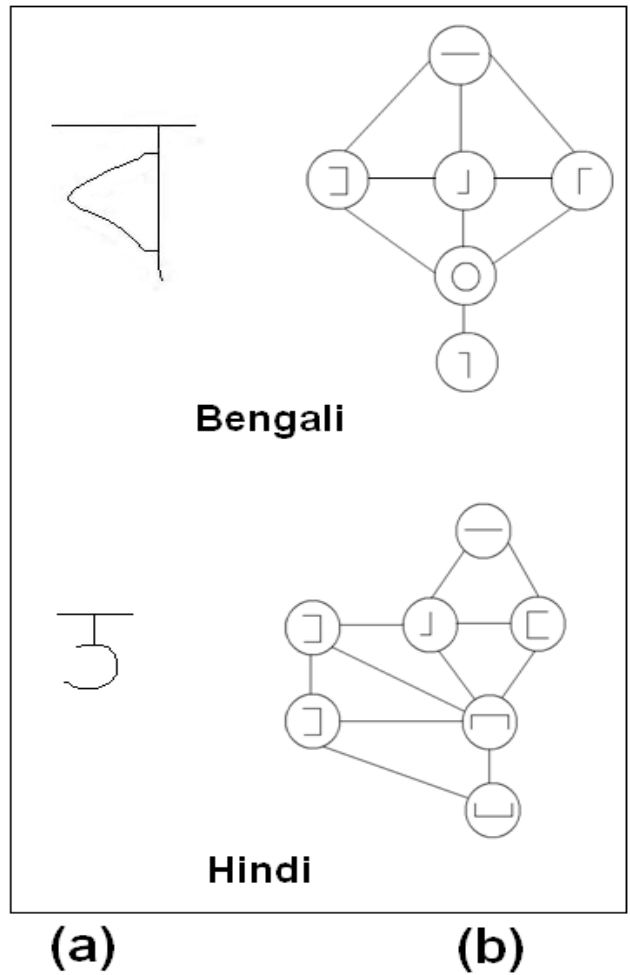

Figure 7. (a) Skeleton image; (b) Shape-based graph.

## 3. EXPERIMENTAL RESULTS

We collected characters from several heterogeneous printed and handwritten documents of Bengali and Hindi. The number of characters in the testing set is 2550 (1300 for Bengali and 1250 for Hindi). All the characters are collected in a systematic manner from printed and handwritten pages scanned on a HP scanjet 5590 scanner at 300 dpi. At first, images were thinned by Bag and Harit medial-axis based thinning method [28]. This thinning method has the capability to preserve the shape of character images at junction points and end points. Then we tested our proposed feature extraction method on these thinned images to get topographic features. Figure 8 and 9 show the shape-graph representation of topographic features of printed Bengali and Hindi character image. The algorithm is implemented in C++ programming language using OpenCV 2.0 on Unix/Linux platform.

We applied this proposed method on handwritten Bengali and Hindi character images. Figure 10 shows few results. It is observed that the performance is satisfactory for handwritten characters also. The proper recognition of handwritten character is still a big challenge.

This new feature set can discriminate characters those are very similar in shape. Figure 11 shows few Hindi characters very similar in shape. After extracting topographic features and generating shape-based graph using them, it is shown that these characters are distinguishable with respect to their topographic feature set. Form this point of view; we can conclude that this new approach will help to improve the accuracy rate of existing Bengali and Hindi OCR systems.

In future, we shall use these topographic features for optical character recognition for handwritten and printed. The brief outline of the recognition method is given here. To identify different topographic features, we give a unique number to each shape. Suppose, *S*00 is marked as 1, *S*01 is as 2, and so on. Now we build a one dimensional feature set, containing different *shape-id*s corresponding to each shape of a particular character. Repeating the same procedure, we prepare a training feature set containing topographic features of different character images. Now for recognition, we apply the same procedure on test sample to extract the features and match the *shape-id*s for checking its presence in the training database.

## 4. CONCLUSION

In this paper, we have proposed a novel topographic feature extraction method for Indian OCR systems. This feature set captures close region, convexity of stroke from different direction, and flat region of thinned character images. The extracted features are represented as a shape-based graph and acts as an invariant feature set for discriminating similar type character images. The proposed method is tested on printed and handwritten Bengali and Hindi characters and we have obtained promising results. In future, we shall extend our work to extract topographic features for other popular Indian languages and make it as a script independent feature set for designing multi-lingual OCR in Indian scripts.

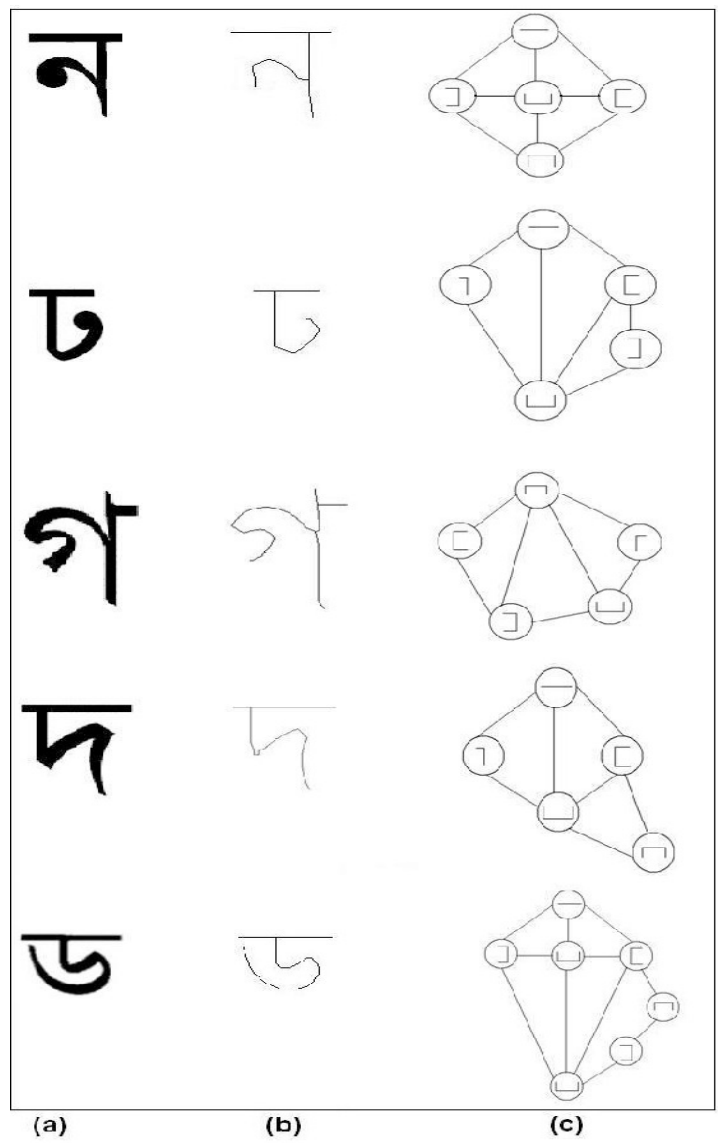

Figure 8. Topographic features of printed Bengali character images: (a) Input image; (b)Skeleton image; (c) Shape-based graph.

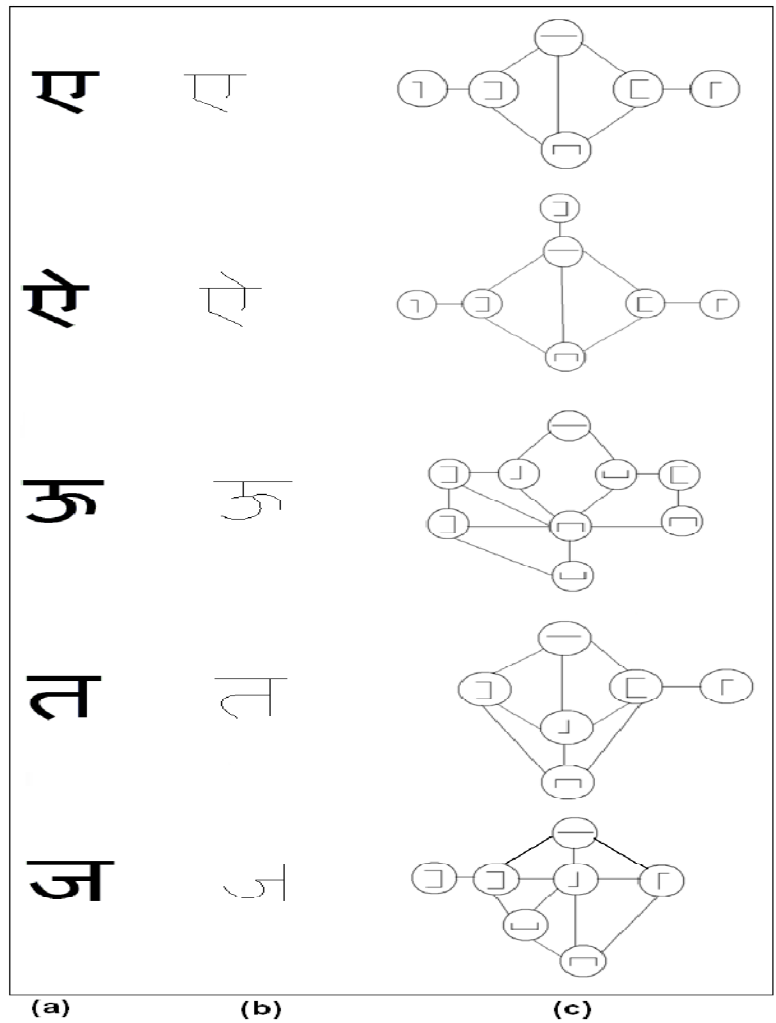

Figure 9. Topographic features of printed Hindi character images: (a) Input image; (b) Skeleton image; (c) Shape-based graph

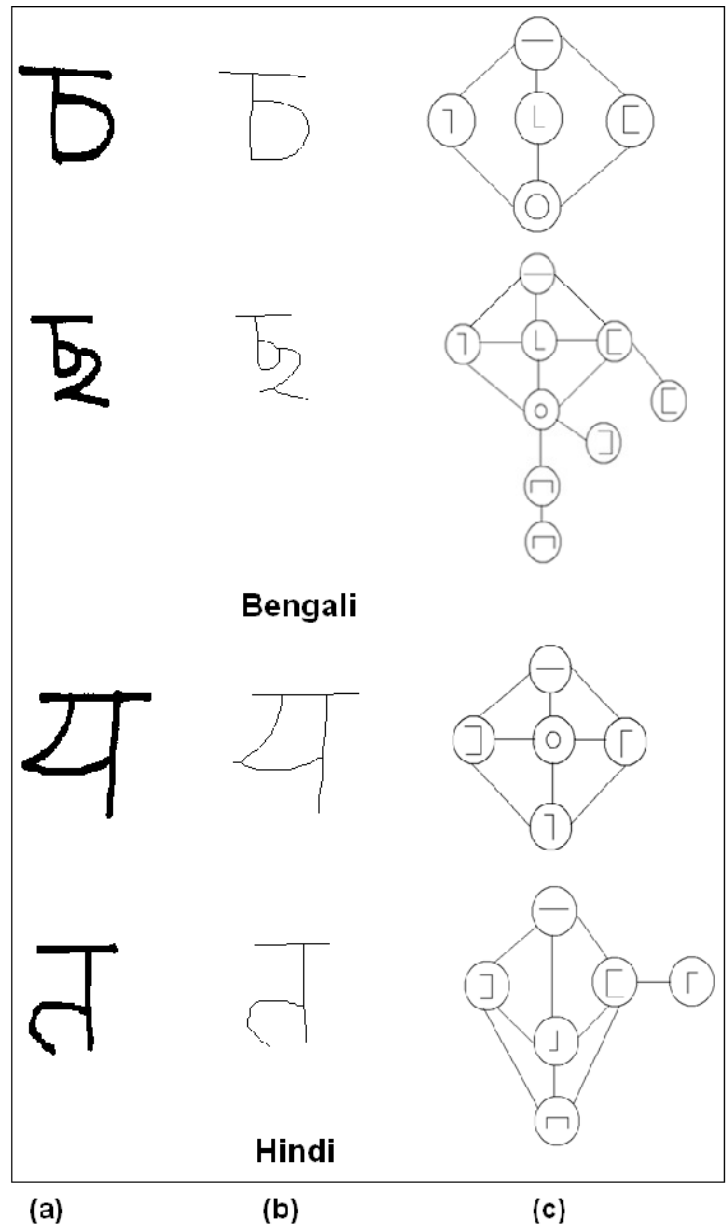

Figure 10. Topographic feature extraction of handwritten character images: (a) Input image; (b) Skeleton image; (b) Shape-based graph.

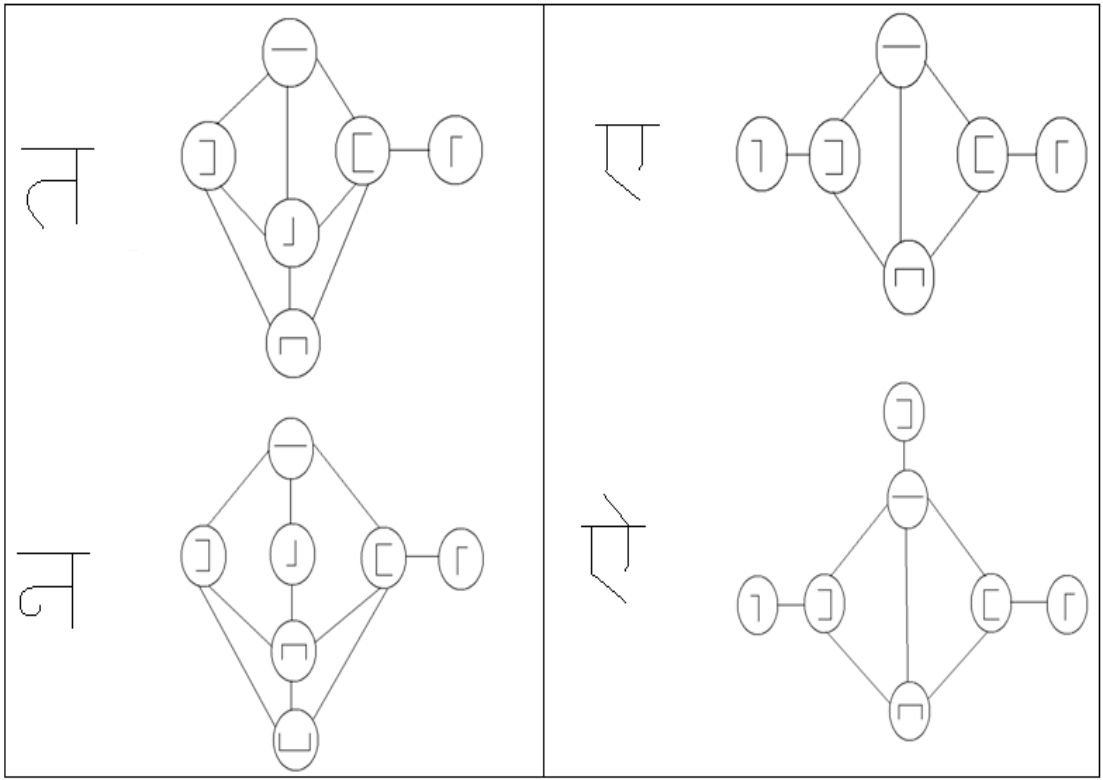

Figure 11. Discrimination between Hindi characters (similar in shape) using shape-based graphs of their topographic features.

…